\documentclass[12pt]{article}
\usepackage{amsmath,amssymb}
\usepackage{graphicx,psfrag,epsf}
\usepackage{enumerate}
\usepackage{natbib}
\usepackage{url} 
\usepackage{hyphenat}
\usepackage{color}

\newcommand{\blind}{0}

\begin{document}

\def\spacingset#1{\renewcommand{\baselinestretch}%
{#1}\small\normalsize} \spacingset{1}

\hyphenpenalty=8000
\tolerance = 100


\def\mytitle{Pruning Variable Selection Ensembles}

\if0\blind
{
  \title{\bf \mytitle}
  \author{Chunxia Zhang\thanks{CZ gratefully acknowledges research support from the 973 Program of China No.~2013CB329406; National Natural Science Foundation of China, No.~61572393 and No.~11671317; and the China Scholarship Council.}\\
    {\normalsize School of Mathematics and Statistics, Xi'an Jiaotong University, China}\\ [0.2cm]
    Yilei Wu and Mu Zhu\thanks{YW and MZ are partially supported by the Natural Sciences and Engineering Research Council of Canada, No.~RGPIN-250419 and No.~RGPIN-2016-03876.}\\
    {\normalsize Department of Statistics and Actuarial Science, University of Waterloo, Canada}}
  \maketitle
} \fi

\if1\blind
{
  \bigskip
  \bigskip
  \bigskip
  \begin{center}
    {\LARGE\bf \mytitle}
\end{center}
  \medskip
} \fi

\bigskip
\begin{abstract}
In the context of variable selection, ensemble learning has gained increasing interest due to its great potential to improve selection accuracy and to reduce false discovery rate. A novel ordering-based selective ensemble learning strategy is designed in this paper to obtain smaller but more accurate ensembles. In particular, a greedy sorting strategy is proposed to rearrange the order by which the members are included into the integration process. Through stopping the fusion process early, a smaller subensemble with higher selection accuracy can be obtained. More importantly, the sequential inclusion criterion reveals the fundamental strength-diversity trade-off among ensemble members. By taking stability selection (abbreviated as StabSel) as an example, some experiments are conducted with both simulated and real-world data to examine the performance of the novel algorithm. Experimental results demonstrate that pruned StabSel generally achieves higher selection accuracy and lower false discovery rates than StabSel and several other benchmark methods.
\end{abstract}

\noindent%
{\it Keywords:} High-dimensional data; Stability selection; Ensemble pruning; Selection accuracy; False discovery rate.
\vfill

\newpage
\spacingset{1.45} 
\section{Introduction}
\label{sec:intro}

With large availability of high-dimensional data in many disciplines, linear regression models play a pivotal role in the analysis due to their simplicity but good performance. In such situations, it is often assumed that the true model is sparse in the sense that only a few covariates have actual influence on the response. Therefore, variable selection is particularly important to detect these covariates to enhance estimation and prediction accuracy, or to improve the interpretability of the model. In this article, we primarily focus on the variable selection problem in a linear regression model,
\begin{equation}
\label{eq:linmodel}
{\bf y}={\bf x}_1\beta_1+{\bf x}_2\beta_2+\cdots+{\bf x}_p\beta_p+\boldsymbol{\varepsilon}={\bf X}\boldsymbol{\beta}+\boldsymbol{\varepsilon},
\end{equation}
where ${\bf y}=(y_1,y_2,\cdots,y_n)^{\rm T}\in\mathbb{R}^n$ is the response vector, ${\bf X}=({\bf x}_1,{\bf x}_2,\cdots,$ ${\bf x}_p)\in\mathbb{R}^{n\times p}$ is the design matrix, and $\{(y_i,{\bf x}_i)\}_{i=1}^n$ are $n$ independent observations. Moreover, $\boldsymbol{\beta}=(\beta_1,\beta_2,\cdots,\beta_p)^{\rm T}\in\mathbb{R}^p$ is a $p$-dimensional unknown coefficient vector and $\boldsymbol{\varepsilon}=(\varepsilon_1,\varepsilon_2,\cdots,\varepsilon_n)^{\rm T}$ $\in\mathbb{R}^n$ is a normally distributed error term, namely, $\boldsymbol{\varepsilon}\sim N({\bf 0},\sigma^2{\bf I})$ in which $\sigma$ is unknown. Here, the response and the covariates are assumed to be mean-corrected; there is thus no intercept term in model (\ref{eq:linmodel}).

Variable selection serves two different objectives depending on whether the modelling purpose is for prediction or for interpretation \citep{Liu2014,Nan2014,Sauerbrei2015,Xin2012}. The former aims to seek a parsimonious model so that future data can be well forecast or {\emph{prediction accuracy}} can be maximized. But for the latter, analysts would like to identify truly important variables (i.e., those having actual influence on an outcome) from numerous candidates, or to maximize {\emph{selection accuracy}}. Due to the significant difference between predictive models and explanatory models, the corresponding variable selection approaches are also very different. In the current paper, we will take selection accuracy (i.e., accurate identification of truly important variables) as our main target.

In the literature, a large number of techniques have been developed to tackle variable selection problems under many different circumstances, such as subset selection \citep{Breiman1996,Miller2002}, coefficient shrinkage \citep{Fan2001,Fan2010,Tibshirani1996}, variable screening \citep{Fan2008}, Bayesian methods \citep{Narisetty2014}, and so on. In high-dimensional situations, much evidence \citep{Breiman1996,Roberts2014,Sauerbrei2015} has demonstrated that some methods (e.g., subset selection, lasso) are unstable. Here, instability of a method means that small changes in data can lead to much variation of the obtained selection results. If prediction is our final objective, this may not affect the result very much because models including different covariates may have comparable prediction ability. Nevertheless, it is particularly crucial to use a stable method to identify important covariates. Take a biological application as an example, biological experts often expect to get a small but stable set of highly informative variables since they need to invest considerable time and research effort to verify them subsequently. In addition, stable results are more reliable and easier to explain. To stabilize these techniques, ensemble learning has great potential since averaging over a number of independent measures is often beneficial.

Ensemble learning, a widely used and efficient technique to enhance the performance of a single learning machine (often called {\emph{base learner}}), has had significant success in solving a large variety of tasks \citep{Kuncheva2014,Schapire2012,Zhou2012}. The main idea of ensemble learning is to make use of the complementarity of many base machines to better cope with a problem. With regard to most existing ensemble methods (e.g., bagging and boosting), they are developed to improve prediction, and the final models obtained can be called {\emph{prediction ensembles}} (PEs). But for {\emph{variable selection ensembles}} (VSEs), a phrase first coined by \citet{Xin2012}, their aim is an accurate identification of covariates which are truly relevant to the response. 

Existing VSE algorithms include: parallel genetic algorithm (PGA) \citep{Zhu2006}, stability selection (StabSel) \citep{Meinshausen2010}, random lasso \citep{Wang2011}, bagged stepwise search (BSS) \citep{Zhu2011}, stochastic stepwise ensembles (ST2E) \citep{Xin2012}, and bootstrap-based tilted correlation screening learning algorithm (TCSL) \citep{Lin2014}. These VSE algorithms usually combine all members to generate an importance measure for each variable. As is the case for PEs, a good strength-diversity trade-off among ensemble members is crucial to the success of a VSE \citep{Zhu2011,Xin2012}. However, there inevitably exist some redundant members which are highly correlated because by definition each member is trying to extract the same information from the same training data. In order to filter out these members to attain better selection results, we propose a novel ordering-based selective ensemble learning strategy to construct more accurate VSEs. The core idea is to first sort ensemble members according to how much they decrease overall {\emph{variable selection loss}} and then  aggregate only those ranked ahead. In particular, the ordering phase is executed by sequentially including the members into an initially empty ensemble so that the variable selection loss of the evolving ensemble is minimized at each step. Then, only those top-ranked members (typically fewer than half of the raw ensemble) are retained to create a smaller but better ensemble. By taking StabSel as an example, our experiments carried out with both simulated and real-world data illustrate that the pruned ensemble does indeed exhibit better performance --- in terms of both selecting the true model more often and reducing the false discovery rate --- than the original, full ensemble as well as many other benchmark methods.


The remainder of this paper is organized as follows. Section 2 presents some related works about VSEs and selective ensemble learning. Section 3 is devoted to proposing a novel, sequential, ordering-based ensemble pruning strategy. Some theoretical insights are also offered for the sequential inclusion criterion, which is shown to balance the strength-diversity trade-off among ensemble members. Section 4 explains how to apply our pruning strategy to StabSel. Some experiments are conducted with a batch of simulations and two real-world examples to evaluate the performance of our proposed methodology, in Sections 5 and 6, respectively. Finally, Section 7 offers some concluding remarks.

\section{Related Works}
\label{section:2}

In this section, we review some related works and ideas. First, there is the basic idea of VSEs.
Like PEs, the process of creating a VSE can be generally divided into two steps, that is, {\it ensemble generation} and {\it ensemble integration}. Most, if not all, existing VSEs utilize a simple averaging rule to assign each variable a final importance measure. Therefore, the key difference among them lies in how to produce a collection of accurate but diverse constituent members. The usual practice is either to execute a base learner (i.e., a variable selection algorithm) on slightly different data sets or to inject some randomness into the base learning algorithm. Among methods of the first type, researchers generally perform selection on a series of bootstrap samples. The representatives include StabSel \citep{Meinshausen2010}, random lasso \citep{Wang2011}, BSS \citep{Zhu2011} and TCSL \citep{Lin2014}. More recently, \citet{Bin2016} systematically studied the efficiency of subsampling and bootstrapping to stabilize forward selection and backward elimination. The core idea behind methods of the second type is to use a stochastic rather than a deterministic search algorithm to perform variable selection. Approaches such as PGA \citep{Zhu2006} and ST2E \citep{Xin2012} belong to this class.

Next, we would like to discuss StabSel in more detail, as we will use it as the main example to illustrate how our method works. StabSel \citep{Meinshausen2010} is a general method that combines subsampling with a variable selection algorithm. Its main idea is to first estimate the selection frequencies of each variable by repeatedly applying a variable selection method to a series of subsamples. Afterwards, only the variables whose selection frequencies exceed a certain threshold are deemed as important. More importantly, the threshold can be chosen in such a way that the expected number of false discoveries can be theoretically bounded under mild conditions. Due to its flexibility and versatility, StabSel has received increasing popularity and been successfully applied in many domains \citep{Beinrucker2016,He2016,Hofner2015,Lin2016,Shah2013} since its inception.


Finally, the idea of {\emph{selective ensemble learning}} (also known as \emph{ensemble pruning}) is not new in  machine learning, either. \citet{Zhou2002} first proved a so-called ``many-could-be-better-than-all'' theorem, which states that it is usually beneficial to select only some, instead of keeping all, members of a learning ensemble. Since then, a great number of ensemble pruning techniques \citep{Chung2015,Guo2013,Hernandez2011,Martinez2009,Mendes2012} have been proposed. Compared with full ensembles, not only are pruned ensembles more efficient both in terms of storage and in terms of prediction speed, they typically also achieve higher prediction accuracy, a win-win situation. Among existing methods, ranking-based and search-based strategies are the two most widely used approaches to select the optimal ensemble subset. The former works by first ranking all the individuals according to a certain criterion and then keeping only a small proportion of top-ranked individuals to form a smaller subensemble. With respect to the latter, a heuristic search is generally carried out in the space of all possible ensemble subsets by evaluating the collective strength of a number of candidate subsets. However, these ensemble pruning methods are all devised for PEs. Due to the significant difference between PEs and VSEs, they cannot be directly used for VSEs.

In the literature of VSEs, only \citet{Zhang2017} had made an attempt to prune VSEs to the best of our knowledge. Nevertheless, our proposed method differs from theirs in several aspects. First, the strategy to sort ensemble members is quite different. Given a VSE, \citet{Zhang2017} used prediction error to evaluate and sort its members, whereas our algorithm sequentially looks for optimal subensembles to minimize variable selection loss. Second, their experiments showed that their subensembles typically do not perform well until they reach at least a certain size; a small ensemble formed by just a few of their top-ranked members usually does not work as well as a random ensemble of the same size. Our method does not suffer from this drawback; empirical experiments show that our subensembles almost always outperform the full ensemble, regardless of their sizes. Last but not least, while \citet{Zhang2017} focused on pruning VSEs from the ST2E algorithm \citep{Xin2012}, in this paper we will primarily focus on applying our pruning technique to stability selection \citep{Meinshausen2010}, for which some extra tricks are needed since stability selection does not aggregate information from individual members by simple averaging.

\section{An Ordering-Based Pruning Algorithm for VSEs}
\label{section:3}

Roughly speaking, the working mechanism of almost all VSEs can be summarized as follows. Each base machine first estimates whether a variable is important or not. By averaging the outputs of many base machines, the ensemble generates an average importance measure for each candidate variable. Subsequently, the variables are ranked according to the estimated importance to the response. To determine which variables are important, a thresholding rule can be further implemented. Essentially, the output of each ensemble member can be summarized by an importance vector with each element reflecting how important the corresponding variable is to the response, and so can the output of the ensemble. Suppose that the true importance vector exists; then, each given member and the overall ensemble will both incur some loss due to their respective departures from the true vector. In this way, given a VSE an optimal subset of its members can be found to minimize this {\emph{variable selection loss}}.

To state the problem formally, we first introduce some notations. Let ${\bf r}^{\ast}=(r_1^{\ast},r_2^{\ast},\cdots,r_p^{\ast})^{\rm T}$ $(\sum_{j=1}^pr_j^{\ast}=1, r_j^{\ast}\geq0)$ denote the {\emph{true}} importance vector, which is not available in practice. The matrix ${\bf R}$ stores the {\emph{estimated}} importance measures, say, ${\bf r}_b=({r}_{b1},{r}_{b2},\cdots,{r}_{bp})^{\rm T}$ $(\sum_{j=1}^pr_{bj}=1, r_{bj}\geq0)$ $(b=1,2,\cdots,B)$, which are produced by $B$ ensemble members. Let $L({\bf r}^{\ast},{\bf r}_b)$ denote the variable selection loss of member $b$. In this paper, we adopt the commonly used squared loss function to measure the loss, i.e., $L({\bf r}^{\ast},{\bf r}_b) = \sum_{j=1}^p(r_j^{\ast} - {r}_{bj})^2=||{\bf r}^{\ast} - {\bf r}_b||_2^2$. 

Then, the loss function for an ensemble of size $B$ is
\begin{equation}
\left\|\frac{1}{B}\sum_{b=1}^B{\bf r}_b - {\bf r}^{\ast}\right\|^2=\frac{1}{B^2}\left\|\sum_{b=1}^B({\bf r}_b - {\bf r}^{\ast})\right\|^2. \label{eq:altloss}
\end{equation}
If we
define a matrix ${\bf E}$ with its element ${\bf E}_{ij}\ (i,j=1,2,\cdots,B)$ as
\begin{equation}
\label{eq:eij}
{\bf E}_{ij} = ({\bf r}_i - {\bf r}^{\ast})^{\rm T}({\bf r}_j - {\bf r}^{\ast}),
\end{equation}
the loss function in (\ref{eq:altloss}) can thus be expressed as
$$
\frac{1}{B^2}\sum_{i=1}^B\sum_{j=1}^B{\bf E}_{ij}.
$$
Assume that there exists a subensemble which can achieve lower loss than the full ensemble, the process of finding the optimal subset is NP-hard, as there are altogether $2^B-1$ non-trivial candidate subsets. Despite this, we can still design an efficient, greedy algorithm that sequentially looks for the best local solution at each step.

To prune the original VSE, we try to sequentially select a subensemble composed of $U<B$ individuals $\{s_1,s_2,\cdots,s_U\}$ that minimizes the loss function
\begin{equation*}
\frac{1}{U^2}\sum_{i=1}^U\sum_{j=1}^U{\bf E}_{s_is_j}.
\end{equation*}
First, the member which has the lowest value of ${\bf E}_{ii}\ (i=1,2,\cdots,B)$ is chosen. In each subsequent step, each candidate member is included into the current ensemble and the one which minimizes the loss is selected. Repeat this process until there is no candidate member left, and we can get a new aggregation order of all the ensemble members. The time complexity of this operation is of polynomial order $O(B^2 p+B^3)$ while that of the exhaustive search is of exponential order. Algorithm 1 lists the main steps of this greedy method to sort the ensemble members of a VSE.

\begin{center}
{\small{\bf Algorithm 1}. A greedy pruning algorithm for VSEs.}
\end{center}
\vskip -0.8cm
\centerline{\makebox[\textwidth]{\hrulefill}}
{\small{\bf{Input}}\vskip 0.2cm
\noindent${\bf R}=({\bf r}_1,{\bf r}_2,\cdots,{\bf r}_B)$: a $p\times B$ matrix storing the importance measures estimated by $B$ members. \vskip 0.1cm
\noindent${\bf r}_{ref}$: a reference importance vector to be used in place of ${\bf r}^{\ast}$ in practice. \vskip 0.2cm
\noindent{\bf{Output}}\vskip 0.1cm
\noindent$\mathcal{S}$: indices for the ordered members.
\vskip 0.2cm
\noindent{\bf{Main steps}}
\begin{enumerate}
\item[1.]{According to formula (\ref{eq:eij}), compute each element ${\bf E}_{ij}$ of the matrix ${\bf E}$ --- replacing ${\bf r}^{\ast}$ with ${\bf r}_{ref}$ in practice.}
\item[2.]{Initialize $\mathcal{S}=\{b\}$, $\mathcal{C}=\{1,2,\cdots,B\}\setminus\{b\}$, where member $b$ is the most accurate one, i.e., $b={\rm argmin}_{1\leq i\leq B}\ {\bf E}_{ii}$.}
\item[3.]{For $u = 2,\cdots, B$}
\begin{enumerate}
\item[(1)]{Let $minumim\longleftarrow +\infty$.}
\item[(2)]{For $k$ in $\mathcal{C}$}
\begin{itemize}
\item{Let
\begin{equation}
\label{eq:algoObjective}
value\longleftarrow \displaystyle\frac{1}{u^2}\left(\sum\limits_{i=1}^{u-1}\sum\limits_{j=1}^{u-1}{\bf E}_{s_is_j} + 2\sum\limits_{i=1}^{u-1}{\bf E}_{s_ik} + {\bf E}_{kk}\right).
\end{equation}}
\item{If ($value < minimum$), let $sel\_{ind}=k$ and $minimum = value$.}
\end{itemize}
\item[]{End For}
\item[(3)]{Let $\mathcal{S}=\mathcal{S}\cup\{sel\_ind\}$ and  $\mathcal{C}=\mathcal{C}\setminus\{sel\_ind\}$.}
\end{enumerate}
\item[]{End For}
\item[4.]{Output the indices of the sorted members $\mathcal{S}$.}
\end{enumerate}}
\vskip -0.8cm
\noindent\makebox[\textwidth]{\hrulefill}

Notice that the true importance vector ${\bf r}^{\ast}$ is generally unknown. Therefore, in practice we need to define a certain reference vector, say, ${\bf r}_{ref}$, to be used as a surrogate of ${\bf r}^{\ast}$. Any rough estimate of ${\bf r}^{\ast}$ can be used. In fact, it is not crucial for ${\bf r}_{ref}$ to be in any sense a ``correct'' importance vector. It is merely used as a rough guide so that we can assess the relatively accuracy of ${\bf r}_b$ (in terms of its partial effect on ${\bf r}_{ref}$); the final variable selection decision is still based upon information contained in $\{{\bf r}_1,...,{\bf r}_B\}$ rather than upon ${\bf r}_{ref}$ itself. In this paper, we use stepwise regression to construct such a reference vector. Given data $({\bf X},{\bf y})$, a linear regression model is estimated by stepwise fitting. Based on the final coefficient estimates $\hat{\beta}_j\ (j=1,2,\cdots,p)$, ${\bf r}_{ref}$ is simply computed as ${\bf r}_{ref}=|\hat{\beta}_j|/\sum_{j=1}^p|\hat{\beta}_j|$.

\def\r{{\bf r}}
Interestingly, one can discern from this sequential algorithm the fundamental ``strength-diversity trade-off'' that drives all ensemble learning algorithms. In Algorithm~1, equation~(\ref{eq:algoObjective}) is the loss after $\r_k$ is added to the current ensemble, $\{\r_{s_1},\r_{s_2},...,\r_{s_{u-1}}\}$. Let
\[
\r_{-k} = \frac{1}{u-1} \sum_{j=1}^{u-1} \r_{s_j}
\]
be the current ensemble estimate, where the subscript ``$-k$'' is used to denote ``prior to having added $\r_k$''. Then, an alternative way to express (\ref{eq:algoObjective}) is
\begin{multline}
\left\|\left[\left(1-\frac{1}{u}\right)\r_{-k}+\frac{1}{u}\r_k\right]-\r^{\ast}\right\|^2 =
\left\|\frac{1}{u}(\r_k-\r_{-k})-(\r^{\ast}-\r_{-k})\right\|^2 \\ =
\frac{1}{u^2}\left\|\r_k-\r_{-k}\right\|^2
-\frac{2}{u}\left(\r_k-\r_{-k}\right)^{\rm T}\left(\r^{\ast}-\r_{-k}\right)
+\left\|\r^{\ast}-\r_{-k}\right\|^2,
\end{multline}
where the last term $\|\r^{\ast}-\r_{-k}\|^2$ is the loss incurred by the ensemble without $\r_k$. Clearly, $\r_k$ can further reduce the overall ensemble loss only if
\[
\frac{2}{u}\left(\r_k-\r_{-k}\right)^{\rm T}\left(\r^{\ast}-\r_{-k}\right)>
\frac{1}{u^2}\left\|\r_k-\r_{-k}\right\|^2
\]
or, equivalently,
\begin{equation}
\label{eq:incl_condition}
\frac{\left(\r_k-\r_{-k}\right)^{\rm T}\left(\r^{\ast}-\r_{-k}\right)}
     {\|\r_k-\r_{-k}\|\|\r^{\ast}-\r_{-k}\|}>
\frac{1}{2u}\times\frac{\left\|\r_k-\r_{-k}\right\|}{\|\r^{\ast}-\r_{-k}\|}.
\end{equation}
The left-hand side of (\ref{eq:incl_condition}) can be viewed as the partial correlation between $\r_k$ and $\r^{\ast}$ after having removed the current estimate $\r_{-k}$ from both; it is thus a measure of how \emph{useful} candidate $\r_k$ is. The right-hand side, on the other hand, is a measure of how \emph{different} candidate $\r_k$ is from the current estimate $\r_{-k}$ (relative to the difference between $\r^{\ast}$ and $\r_{-k}$). Hence, one can interpret condition (\ref{eq:incl_condition}) as a lower bound on the usefulness of $\r_k$ in order for it to be considered a viable candidate as the next (i.e., the $u$-th) member of the ensemble.

First, the bound decreases with the index $u$, that is, the bar of entry is steadily lowered as more and more members are added. This is necessary --- since it is more difficult to improve an already sizable ensemble, a new member becomes admissible as long as it has {\em some} additional value. Second, if the new candidate $\r_k$ is very different from $\r_{-k}$, then it must be very useful as well --- in terms of its partial correlation with $\r^{\ast}$ --- in order to be considered. This observation is consistent with a fundamental trade-off in ensemble learning, referred to as the ``strength-diversity trade-off'' by Leo Breiman in his famous paper on random forests \citep{Breiman2001}, which implies that something very different (diversity) had better be very useful (strength). The analysis above thus provides some crucial insight about how the accuracy and diversity of individuals in a pruned VSE work together to improve its performance.

Based on the output $\mathcal{S}$ of Algorithm 1, we can create the average importance for the $p$ variables by averaging the results of only some --- say, the top $U$ --- members of the full ensemble. Then, the variables can be ordered accordingly. Ideally, the value of $U$ should be automatically determined to maximize selection accuracy. However, variable selection accuracy is not as readily computable as is prediction accuracy, since the truly important variables are unknown in practice. The easiest method is to prescribe a desired number for $U$. According to our experiments (refer to Section 5) as well as evidence in the study of PEs \citep{Martinez2009,Hernandez2011}, it often suffices to keep only the first 1/4 to 1/2 of the sorted members.

\section{Ensemble Pruning for Stability Selection}
\label{section:4}

While the pruning algorithm (Algorithm 1) we provided in Section~\ref{section:3} can be applied to any VSE, in this paper we will use StabSel \citep{Meinshausen2010} as an example to demonstrate its application and effectiveness, in view of StabSel's popularity and flexibility in high-dimensional data analysis \citep[see, e.g.,][]{Beinrucker2016,He2016,Hofner2015,Lin2016,Shah2013}.

In essence, StabSel is an ensemble learning procedure. In the generation stage, it applies the base learner lasso (or randomized lasso) repeatedly to subsamples randomly drawn from the training data. When combining the information provided by the multiple lasso learners, it employs a special strategy, as opposed to simple averaging. For each candidate value of the tuning parameter in the lasso, it first estimates the probability that a variable is identified to be important. Next, it assigns an importance measure to each variable as the maximum probability that the variable is considered as important over the entire regularization region consisting of all candidate tuning parameters. Finally, the selection result is obtained by evaluating the importance measures against a given threshold. To ease presentation, we summarize the main steps of StabSel, with the lasso as its base learner, as Algorithm 2. That StabSel uses an aggregation strategy other than that of simple averaging is another reason why we have chosen to use it as our main example, because it is less obvious how our pruning algorithm should be applied.

\begin{center}
{\small{\bf Algorithm 2}. Stability selection for variable selection.}
\end{center}
\vskip -0.8cm
\centerline{\makebox[\textwidth]{\hrulefill}}
{\small{\bf{Input}}\\
${\bf y}$: a $n\times 1$ response vector. \\
${\bf X}$: a $n\times p$ design matrix. \\
$\Lambda$: a set containing $K$ regularization parameters for the lasso.\\
$B$: number of base learners.\\
$\pi_{\rm thr}$: a pre-specified threshold value.\\
\noindent{\bf{Output}}\\
$\mathcal{I}$: index set of the selected variables.\\
\noindent{\bf{Main steps}}
\begin{enumerate}
\item{For $b=1,2,\cdots,B$}
\begin{itemize}
\item{Randomly draw a subsample $({\bf X}',{\bf y}')$ of size $\lfloor\frac{n}{2}\rfloor$ from $({\bf X},{\bf y})$ where $\lfloor x\rfloor$ denotes the largest integer less than or equal to $x$.}
\item{With each regularization parameter $\lambda_k\in\Lambda$, execute lasso with $({\bf X}',{\bf y}')$ and denote the selection results as $\hat{\mathcal{S}}^{\lambda_k}_{\lfloor n/2\rfloor,b}\ (k=1,2,\cdots,K)$.}
\end{itemize}
\item[]{EndFor}
\item{Compute the selection frequencies for each variable as
\begin{equation}
\hat{\pi}_j=\underset{\lambda_k\in\Lambda}{\max}\displaystyle\left\{\hat{\pi}_j^{\lambda_k}\right\},\quad j=1,2,\cdots,p,
\end{equation}
in which
\begin{equation}
\label{eq:StabSelAgg}
\hat{\pi}_j^{\lambda_k}\triangleq P^{\ast}\{j\in\hat{\mathcal{S}}^{\lambda_k}\}=\frac{1}{B}\sum_{b=1}^B\mathbb{I}\{j\in\hat{\mathcal{S}}^{\lambda_k}_{\lfloor n/2\rfloor,b}\},
\end{equation}
where $\mathbb{I}(\cdot)$ is an indicator function which is equal to 1 if the condition is fulfilled and to 0 otherwise.}
\item{Select the variables whose selection frequency is above the threshold value, i.e., $\mathcal{I}=\{j:\hat{\pi}_j\geq\pi_{\rm thr}\}$.}
\end{enumerate}}
\vskip -0.8cm
\noindent\makebox[\textwidth]{\hrulefill}

Let $V$ stand for the number of variables wrongly selected by StabSel, i.e., false discoveries. Under mild conditions on the base learner, \citet{Meinshausen2010} proved that the expected value of $V$ (also called per-family error rate, PFER) can be bounded for $\pi_{\rm thr}\in(\frac{1}{2},1)$ by
\begin{equation}
\label{eq:bound}
E(V)\leq\frac{1}{2\pi_{\rm thr} - 1}\cdot\frac{q_{\Lambda}^2}{p},
\end{equation}
in which $q_{\Lambda}$ means the number of selected variables per base learner.
%
%
Following guidelines provided by \citet{Meinshausen2010}, we set $\pi_{thr}=0.7$. Furthermore, we choose a false discovery tolerance of $E(V) \leq 4$, which implies a targeted value of $q_{\Lambda}=\lceil(1.6p)^{1/2}\rceil$ variables to be chosen by the base learner according to (\ref{eq:bound}). Hence, we set the regularization region $\Lambda$ for the lasso to be a set of $K$ different values --- equally spaced on the logarithmic scale --- between $\lambda_{\rm min}$ and $\lambda_{\rm max}$, where $\lambda_{\rm max}=\max_j|n^{-1}{\bf x}_j^{\rm T}{\bf y}|$ and
\begin{equation*}
\lambda_{\rm min}=\underset{\lambda}{\rm argmax}\left\{|\lambda_{\rm max}-\lambda|: \ 0\leq\lambda\leq\lambda_{\rm max}, \ q_{\lambda}=\lceil(1.6p)^{1/2}\rceil\right\}.
\end{equation*}
That is, starting from $\lambda_{\rm max}$, we push $\lambda_{\rm min}$ far enough until the lasso is able to include $q_{\Lambda}=\lceil(1.6p)^{1/2}\rceil$ number of variables. In our experiments, we usually took $K=100$.

For StabSel, the selection results of each ensemble member, $\hat{\mathcal{S}}^{\lambda_k}_{\lfloor n/2 \rfloor,b}\ (k=1,2,\cdots,K)$ (see Algorithm 2, Step 1), is a matrix ${\bf T}^{(b)}$ of size $p\times K$, rather than simply an importance vector of length $p$. Each entry ${\bf T}^{(b)}_{jk}\ (j=1,\cdots,p;k=1,\cdots,K)$ is a binary indicator of whether variable $j$ is selected when the regularization parameter is $\lambda_k$. When applying Algorithm 1 to rearrange the aggregation order of the ensemble members in StabSel, each ${\bf T}^{(b)}$ needs to be transformed into an importance vector ${\bf r}_b=(r_{b1},r_{b2},\cdots,r_{bp})^{\rm T}$, with $r_{bj}$ reflecting the importance of variable $j$ as estimated by member $b$. To achieve this, $r_{bj}$ is computed as $r_{bj}=(1/K)\sum_{k=1}^K{\bf T}^{(b)}_{jk}$, or
\begin{equation}
\label{eq:condense}
{\bf r}_b = \frac{1}{K} {\bf T}^{(b)} {\bf 1}.
\end{equation}
That is, a variable is deemed more important if it is selected over a larger regularization region.

Overall, the pruned StabSel algorithm works by inserting our ensemble pruning algorithm (Algorithm 1) as an extra step into the StabSel algorithm (Algorithm 2). First, a StabSel ensemble of size $B$ is generated ``as usual'' by step 1 of Algorithm 2. Then, the selection result ${\bf T}^{(b)}$ of each ensemble member is condensed into ${\bf r}_b$ according to (\ref{eq:condense}) and Algorithm 1 is utilized to sort the members and obtain the ranked list, $\mathcal{S}$. Afterwards, the ensemble members are fused ``as usual'' by steps 2 and 3 of Algorithm 2, except that only the members corresponding to the top $U$ elements of $\mathcal{S}$ are fused rather than all of the original $B$ members --- specifically, only the top $U$ members are used when computing $\hat{\pi}_j^{\lambda_k}$ in (\ref{eq:StabSelAgg}).

\section{Simulation Studies}
\label{section:5}

In this section, some experiments are conducted with simulated data in different experimental settings to study the performance of the pruned StabSel algorithm. Particularly, the effect of modifying the aggregation order of ensemble members in StabSel is first analyzed (scenario 1 below). Then (scenarios 2-5 below), pruned StabSel is examined and compared with vanilla StabSel as well as some other popular benchmark methods including the lasso \citep{Tibshirani1996}, SCAD \citep{Fan2001} and SIS \citep{Fan2008}.

In the following experiments, the lasso and StabSel were implemented by the {\tt glmnet} toolbox \citep{Qian2013} in Matlab, while SCAD and SIS were available as part of the package {\tt ncvreg} \citep{Breheny2011} in R. In SIS, $\lceil n/\log(n)\rceil$ variables having the largest marginal correlation with the response were first selected and SCAD was then followed to identify important ones. Ten-fold cross-validation was used to select the tuning parameters for the lasso and SCAD.

To extensively evaluate the performance of a method, the following five different measures were employed. Let $\boldsymbol{\beta}^{\ast}=(\beta_1^{\ast},\beta_2^{\ast},\cdots,\beta_p^{\ast})^{\rm T}$ be the coefficient vector for the true model $\mathcal{T}$, i.e., $\mathcal{T}=\{j:\beta_{j}^{\ast}\neq 0\}$. To estimate an evaluation metric, we replicated each simulation $M$ times. In the $m$-th replication, denote $\hat{\boldsymbol{\beta}}_m=(\hat{\beta}_{1,m},\hat{\beta}_{2,m},\cdots,\hat{\beta}_{p,m})^{\rm T}$ as the estimated coefficients and $\hat{\mathcal{S}}_{m}=\{j:\hat{\beta}_{j,m}\neq 0\}$ as the identified model. Moreover, let $d_0$ and $(p-d_0)$ indicate the number of truly important and unimportant variables, respectively. Then, we define
\begin{equation}
\label{eq:perf-prob}
\begin{array}{lc}
\bar{p}_1 = \displaystyle\frac{1}{d_0\times M}\left(\sum_{m=1}^M\sum_{j=1}^p\mathbb{I}(\beta_j^{\ast}\neq0 \mbox{ and } \hat{\beta}_{j,m}\neq 0)\right),\\ [0.10cm]
\bar{p}_0 = \displaystyle\frac{1}{(p - d_0)\times M}\left(\sum_{m=1}^M\sum_{j=1}^p\mathbb{I}(\beta_j^{\ast}=0 \mbox{ and } \hat{\beta}_{j,m}\neq 0)\right),
\end{array}
\end{equation}
\begin{equation}
\label{eq:perf-rate}
acc.=\displaystyle\frac{1}{M}\sum_{m=1}^M\mathbb{I}(\hat{\mathcal{S}}_m=\mathcal{T}),\qquad FDR = \displaystyle\frac{1}{M}\sum_{m=1}^M\sum_{j=1}^p\frac{\mathbb{I}(\beta_j^{\ast}=0 \mbox{ and } \hat{\beta}_{j,m}\neq 0)}{\mathbb{I}(\hat{\beta}_{j,m}\neq 0)},
\end{equation}
\begin{equation}
\label{eq:perf-pred}
PErr = \displaystyle\frac{1}{\sigma^2}E[(\hat{y} - {\bf x}^{\rm T}\boldsymbol{\beta}^{\ast})^2]=\frac{1}{M}\sum_{m=1}^M\left[\frac{1}{\sigma^2}(\hat{\boldsymbol{\beta}}_m - \boldsymbol{\beta}^{\ast})^{\rm T}[E({\bf x}{\bf x}^{\rm T})](\hat{\boldsymbol{\beta}}_m - \boldsymbol{\beta}^{\ast})\right].
\end{equation}
In the formulae above, $\mathbb{I}(\cdot)$ represents an indicator function. The $\bar{p}_1$ and $\bar{p}_0$ defined in (\ref{eq:perf-prob}), respectively, correspond to the {\it mean selection probability} for the truly important and unimportant variables --- i.e., true positive and false positive rates, respectively. In general, a good method should simultaneously achieve a $\bar{p}_1$ value close to 1 and a $\bar{p}_0$ value close to 0. The {\it selection accuracy} (abbreviated as $acc.$) in (\ref{eq:perf-rate}) indicates the frequency that an algorithm exactly identifies the true model, and the {\it false discovery rate} (FDR) assesses the capacity of an approach to exclude noise variables. To evaluate the prediction ability of a method, we utilized the {\it relative prediction error} (simply abbreviated as $PErr$), given in (\ref{eq:perf-pred}), by following the practice of \citet{Zou2006}. In particular, a linear regression model was built by using the selected variables. Then, $E({\bf x}{\bf x}^{\rm T})$ and the relative prediction error were estimated with an independent test set composed of 10,000 instances.

\subsection{Simulated data}
\label{section:5.1}

The simulated data in the following scenarios 1-4 were generated by
\begin{equation*}
{\bf y}={\bf x}_1\beta_1+{\bf x}_2\beta_2+\cdots+{\bf x}_p\beta_p+\boldsymbol{\varepsilon}={\bf X}\boldsymbol{\beta}+\boldsymbol{\varepsilon},\quad \boldsymbol{\varepsilon}\sim N({\bf 0},\sigma^2{\bf I}),
\end{equation*}
where $\boldsymbol{\varepsilon}$ is a normally distributed error term with mean zero and variance $\sigma^2$. In scenario 5, we simulated data from a logistic regression model. Although we have focused mostly on linear regression problems in this paper, all of these ideas (i.e., StabSel, SIS, etc) can be generalized easily to other settings, and so can our idea of ensemble pruning. For logistic regression (scenario 5 here and a real-data example later in Section~\ref{section:6}), rather than the relative prediction error defined in (\ref{eq:perf-pred}), we simply used the average misclassification error,
\[
PErr = \frac{1}{M}\sum_{m=1}^M P(\hat{y}_m \neq y),
\]
with $P(\hat{y}_m \neq y)$ being estimated on an independently generated (or held out) test set, to measure prediction capacity.

\paragraph{Scenario 1}
This is a simple scenario taken from \citet{Zhu2006}, which we used primarily to study the effect of our proposed re-ordering of the ensemble members (see Figures~\ref{fig:1} and \ref{fig:2}), rather than to evaluate its performance against various benchmark algorithms. There are $p=20$ variables and $n=40$ observations. Particularly, only variables ${\bf x}_5, {\bf x}_{10}$ and ${\bf x}_{15}$ have actual influence on the response variable ${\bf y}$ and their true coefficients are 1, 2, 3, respectively. The rest of variables are uninformative. As in \citet{Zhu2006}, for the explanatory variables we considered the following 4 variations:
\begin{center}
\begin{tabular}{ll}
{\emph{Variation 1}}: & ${\bf x}_1,{\bf x}_2,\cdots,{\bf x}_{20}\sim N({\bf 0},{\bf I})$; \\
{\emph{Variation 2}}: & ${\bf x}_1,{\bf x}_2,\cdots,{\bf x}_{19}\sim N({\bf 0},{\bf I})$, ${\bf x}_{20}={\bf x}_{5}+0.25{\bf z}$, ${\bf z}\sim N({\bf 0},{\bf I})$; \\
{\emph{Variation 3}}: & ${\bf x}_1,{\bf x}_2,\cdots,{\bf x}_{19}\sim N({\bf 0},{\bf I})$, ${\bf x}_{20}={\bf x}_{10}+0.25{\bf z}$, ${\bf z}\sim N({\bf 0},{\bf I})$; \\
{\emph{Variation 4}}: & ${\bf x}_j={\bf z}+\boldsymbol{\epsilon}_j, j=1,2,\cdots,20$, $\boldsymbol{\epsilon}_j\sim N({\bf 0},{\bf I})$, ${\bf z}\sim N({\bf 0},{\bf I})$.
\end{tabular}
\end{center}
In variation 1, all covariate 1 are independent. In variations 2 and 3, the variable ${\bf x}_{20}$ is highly correlated with ${\bf x}_{5}$ and with ${\bf x}_{10}$, respectively, each with correlation coefficient $\rho \approx 0.97$. In variation 4, all variables are moderately correlated with each other, with $\rho\approx0.5$. As for the standard deviation $\sigma$ of $\boldsymbol{\varepsilon}$, it was set to be $\sigma=1$ for variations 1-3 and $\sigma=2$ for variation 4. 

\paragraph{Scenario 2}
This is a scenario similar to one considered by \citet{Narisetty2014}.
In this scenario, the covariates come from a normal distribution with a compound symmetric covariance matrix ${\bf \Sigma}=(\Sigma_{i,j})_{p\times p}$ in which $\Sigma_{i,j}=\rho$ for $i\neq j$. Five covariates are truly important to the response and their coefficients are taken as $\boldsymbol{\beta}_{IV}=(0.5, 1.0, 1.5, 2.0, 2.5)^{\rm T}$. The rest of the covariates are considered as unimportant and their coefficients are all zero. By varying the value of $n$, $p$ and $\rho$, we examined the performance of each method for $n\geq p$ and for $n\ll p$ (see Table~\ref{tab:1}).

\paragraph{Scenario 3}
Here, we considered a setting in which the covariates have a block covariance structure, similar to one used by \citet{Narisetty2014}, again. In the model, the signal variables have pairwise correlation $\rho_1=0.25$; the noise variables have pairwise correlation $\rho_2=0.75$; and each pair of signal and noise variables has a correlation of $\rho_3=0.50$. The true coefficient vector is $\boldsymbol{\beta}=(0.5, 1.0, 1.5, 2.0, 2.5,{\bf 0}_{p - 5})^{\rm T}$. We focused our attention on a high-dimensional $p > n$ setting, with $n=200, p = 1000$ and $\sigma=1$.

\paragraph{Scenario 4}
In this scenario, we studied a more challenging problem based on a commonly used benchmark data set \citep{Lin2014,Wang2011,Fan2001,Zou2006,Xin2012}. Here, ${\bf X}$ is generated from a multivariate normal distribution with mean zero and covariance matrix ${\bf \Sigma}=(\Sigma_{i,j})_{p\times p}$, where $\Sigma_{i,j}=\rho^{|i-j|}$ for $i\neq j$. The true coefficient vector is $\boldsymbol{\beta}=(3,1.5,0,0,2,0.5,0.5,{\bf 0}_{p-7})^{\rm T}$ and $\sigma$ is set to 1. Again, we set $n=200$ and $p=1000$ to focus on the $p > n$ case, and took $\rho=0.50, 0.90$ to evaluate the performance of each method. Other than the high correlations, what makes this scenario especially challenging is the existence of two weak signals with true coefficients equal to $0.5$.

\paragraph{Scenario 5}
Finally, we considered a logistic regression model \citep{Fan2001} with data created from
\begin{equation*}
{\rm logit}(p_i)=\log\left(\frac{p_i}{1-p_i}\right)={\bf x}_i^{\rm T}\boldsymbol{\beta},
\end{equation*}
where $p_i$ is the probability that the response variable $y_i$ is equal to $1$. The true coefficient vector is $\boldsymbol{\beta}=(3, 1.5, 0, 0, 2,{\bf 0}_{p - 5})^{\rm T}$. The components of ${\bf x}$ are standard normal, where the correlation between ${\bf x}_i$ and ${\bf x}_j$ is $\rho({\bf x}_i,{\bf x}_j)=0.5^{|i-j|} \ \forall\ i\neq j$. We took $n=200$ and examined both a relatively low-dimensional setting with $p=50$ and a high-dimensional one with $p=1000$.

\subsection{Effect of changing the aggregation order}
\label{section:5.2}

As we stated earlier, we used scenario 1 primarily to analyze our proposed reordering algorithm rather than to make general performance comparison. First, we used it to investigate how the performance of StabSel varies if the aggregation order of its constituent members is rearranged. To evaluate the performance of a VSE, the estimated selection accuracy over 100 simulations was employed. Since the dimensionality $p=20$ is relatively low in scenario 1, we used a slightly different set of parameters --- specifically, $q_{\Lambda}=\lceil0.8p\rceil$ and $\pi_{\rm thr}=0.6$ --- to run StabSel than what we recommended earlier in Section~\ref{section:4} for high-dimensional problems.
\begin{figure}[!t]
\centering
\begin{minipage}[b]{0.49\textwidth}
\centering{\scalebox{0.55}[0.50]{\includegraphics{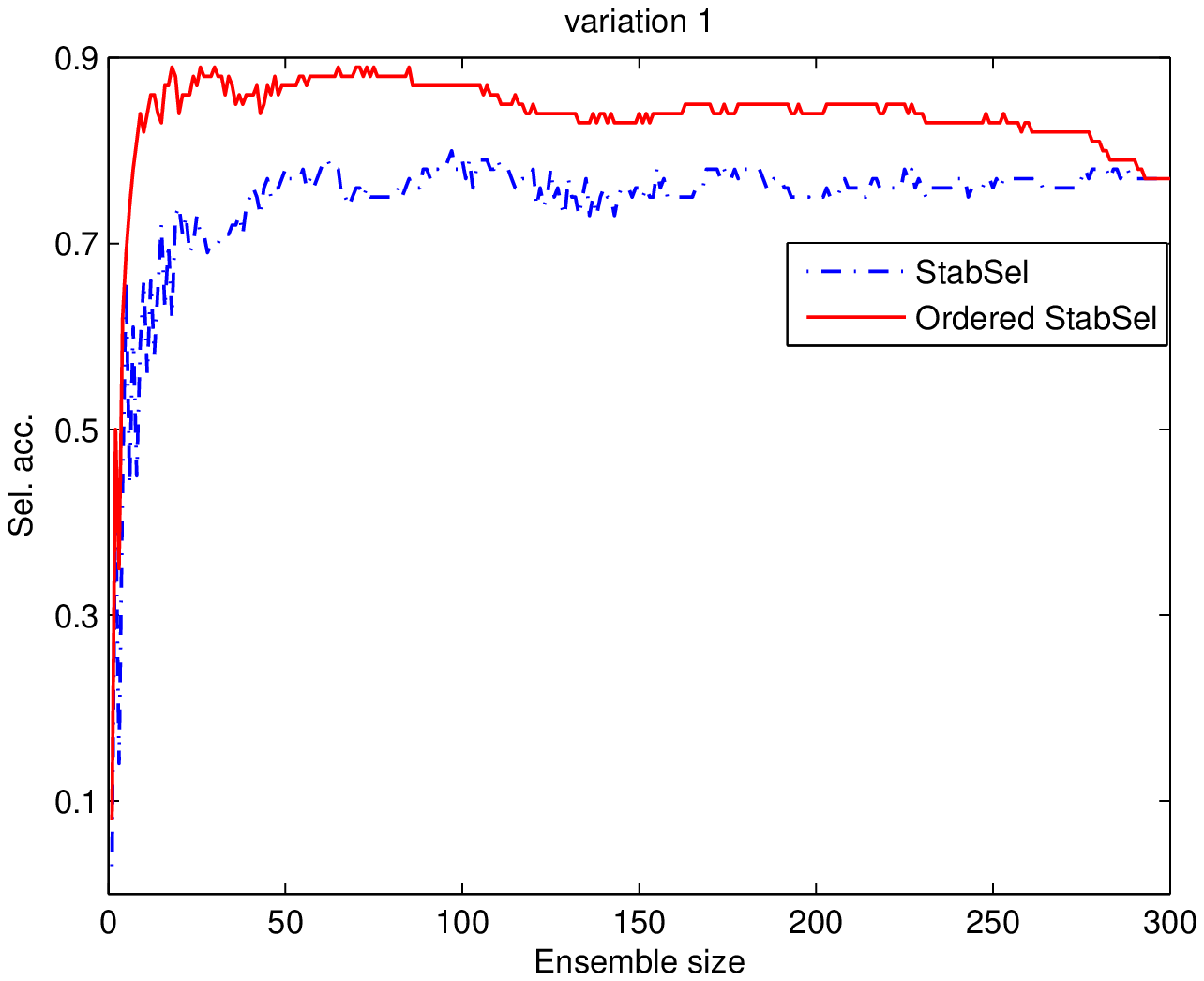}}}
\end{minipage}%
\hspace{0.005\textwidth}%
\begin{minipage}[b]{0.49\textwidth}
{\scalebox{0.55}[0.50]{\includegraphics{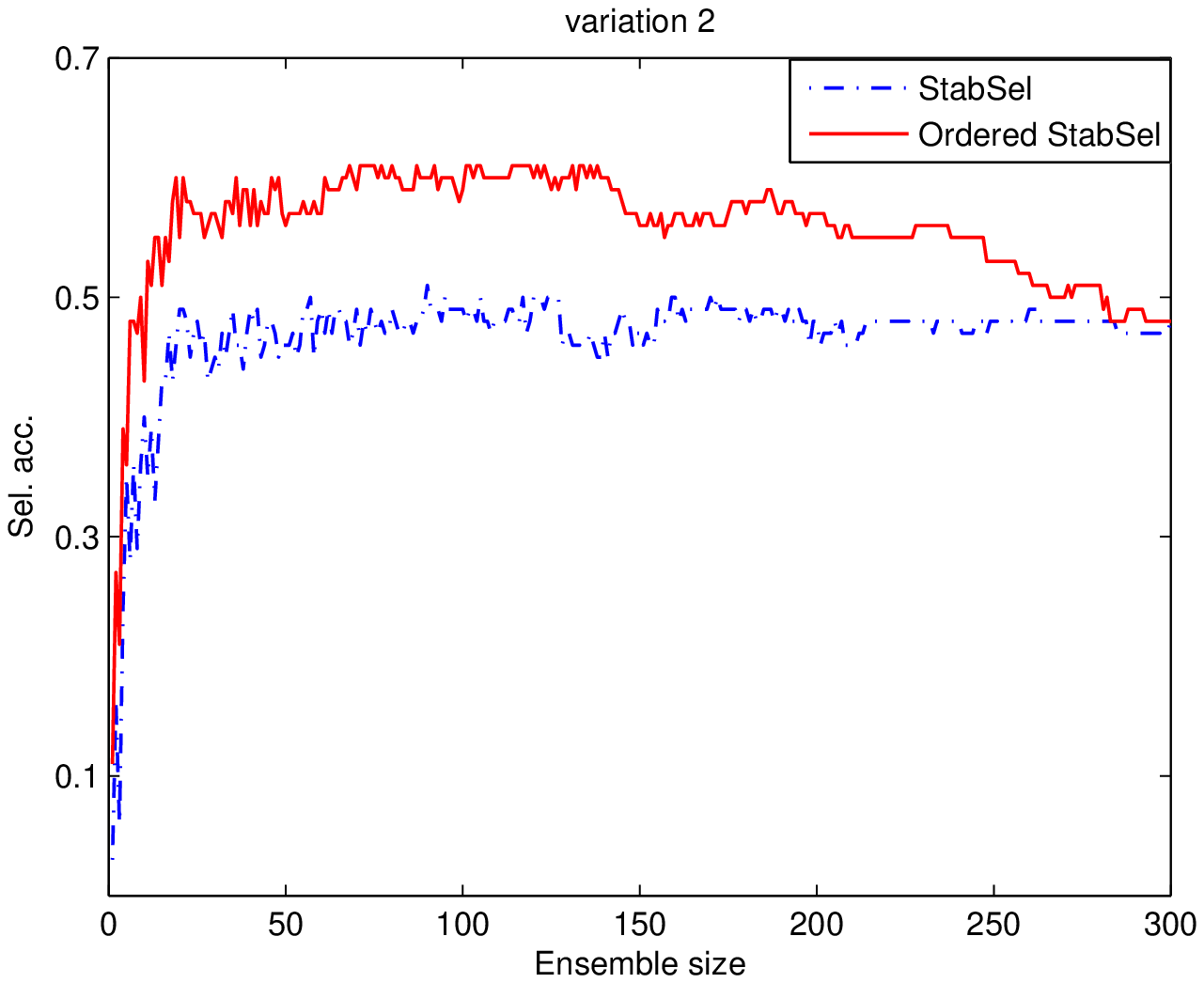}}}
\end{minipage}\\[2pt]
\begin{minipage}[b]{0.49\textwidth}
\centering{\scalebox{0.55}[0.50]{\includegraphics{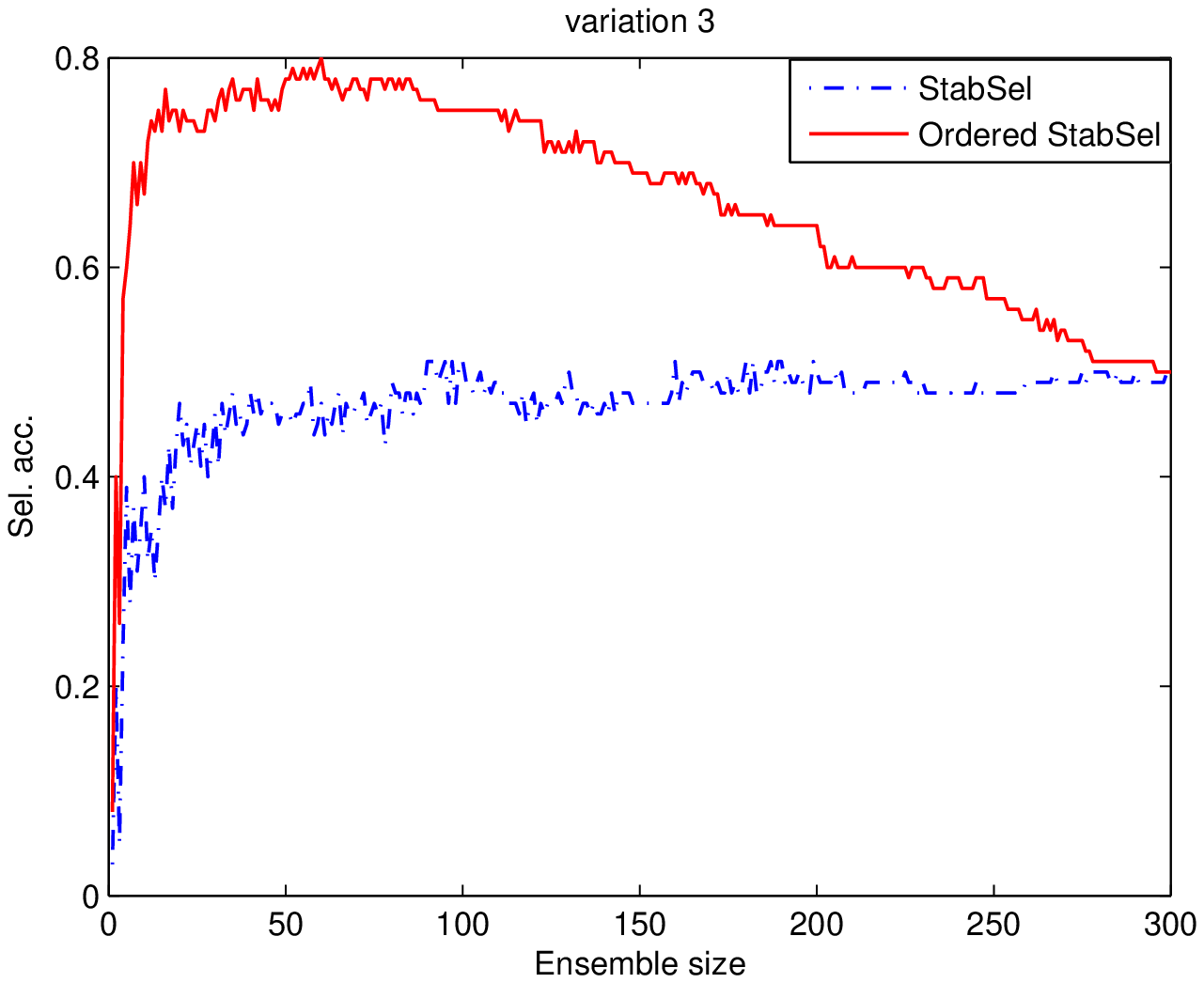}}}
\end{minipage}%
\hspace{0.005\textwidth}%
\begin{minipage}[b]{0.49\textwidth}
{\scalebox{0.55}[0.50]{\includegraphics{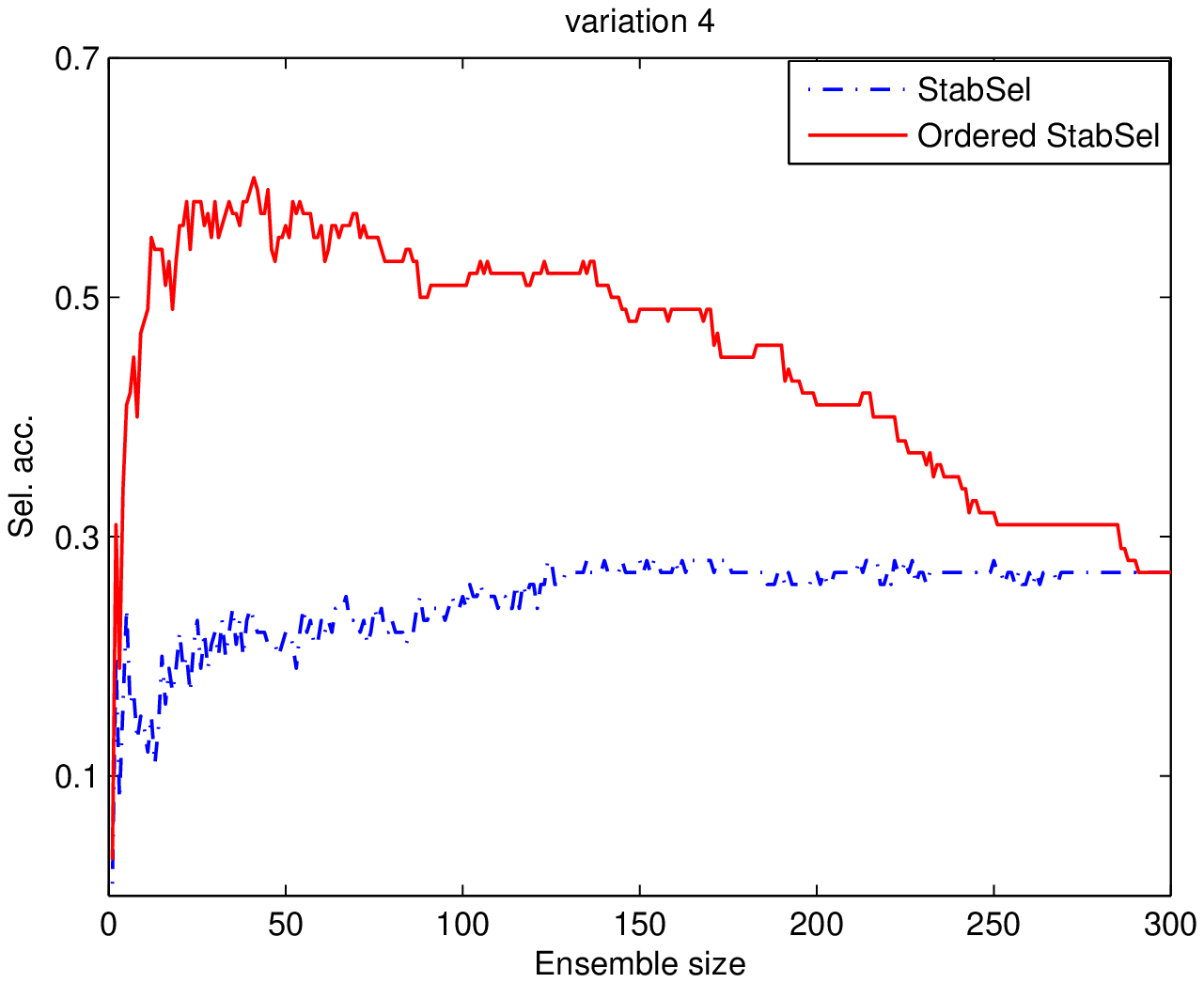}}}
\end{minipage}\\
\caption{The performance of StabSel and its re-ordered version as a function of ensemble size. \label{fig:1}}
\end{figure}

Figure~\ref{fig:1} depicts the selection accuracy for subensembles of (regular) StabSel and of ordered StabSel as a function of their respective sizes. For (regular) StabSel, the members were aggregated gradually in the same random order as they were generated, whereas, for ordered StabSel, the members were first sorted by Algorithm 1 and then fused one by one. It can be observed from Figure 1 that the accuracy of (regular) StabSel tends to increase rapidly at the beginning as more members are aggregated. Then, it quickly reaches a nearly optimal value, after which further improvement by adding more members becomes negligible. But for ordered StabSel, this accuracy curve always reaches a maximum somewhere in the middle; afterwards, the curve steadily declines until it reaches the same level of accuracy as that of (regular) StabSel, when the two algorithms fuse exactly the same set of members. Moreover, we can see that the selection accuracy of almost {\em any} ordered subensemble would be higher than that of the full ensemble consisting of all $B=300$ members (i.e, the rightmost point in each subplot). This unequivocally demonstrates the value of our ordering-based ensemble pruning and selective fusion algorithm. 

\subsection{Effect of the original ensemble size $B$}
\label{section:5.3}

\begin{figure}[!t]
\begin{minipage}[b]{0.49\textwidth}
\centering{\scalebox{0.55}[0.50]{\includegraphics{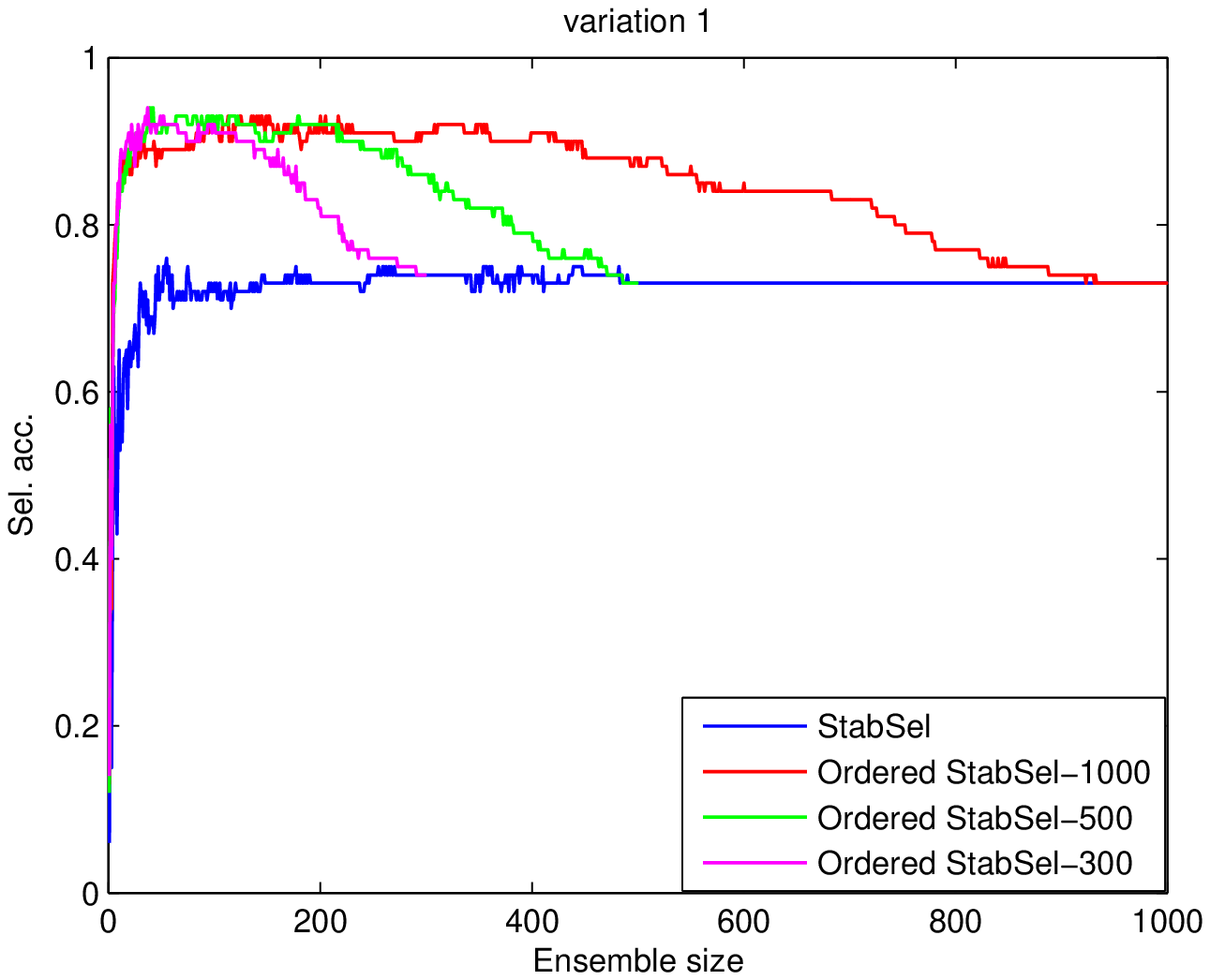}}}
\end{minipage}%
\hspace{0.005\textwidth}%
\begin{minipage}[b]{0.49\textwidth}
{\scalebox{0.55}[0.50]{\includegraphics{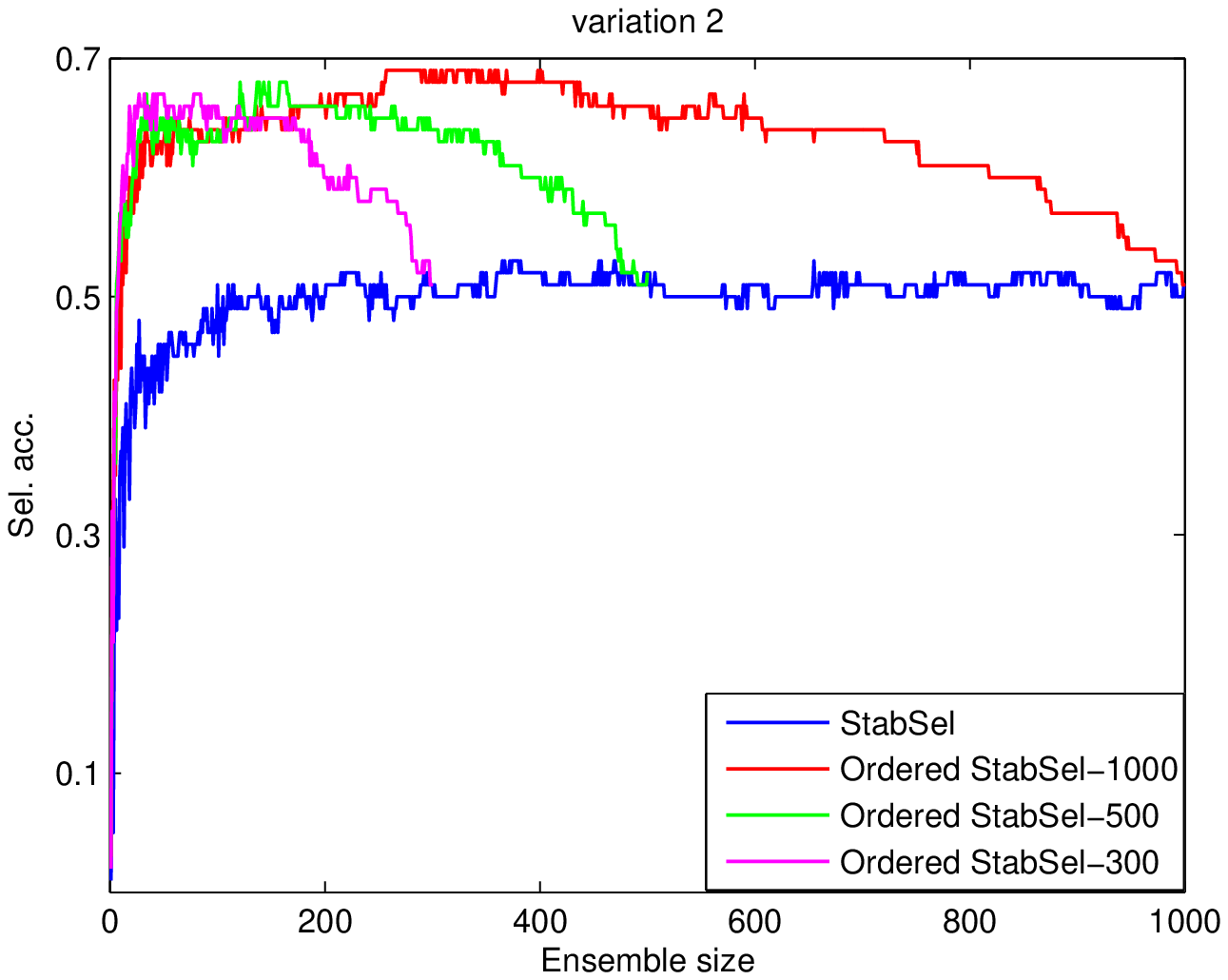}}}
\end{minipage}\\[2pt]
\begin{minipage}[b]{0.49\textwidth}
\centering{\scalebox{0.55}[0.50]{\includegraphics{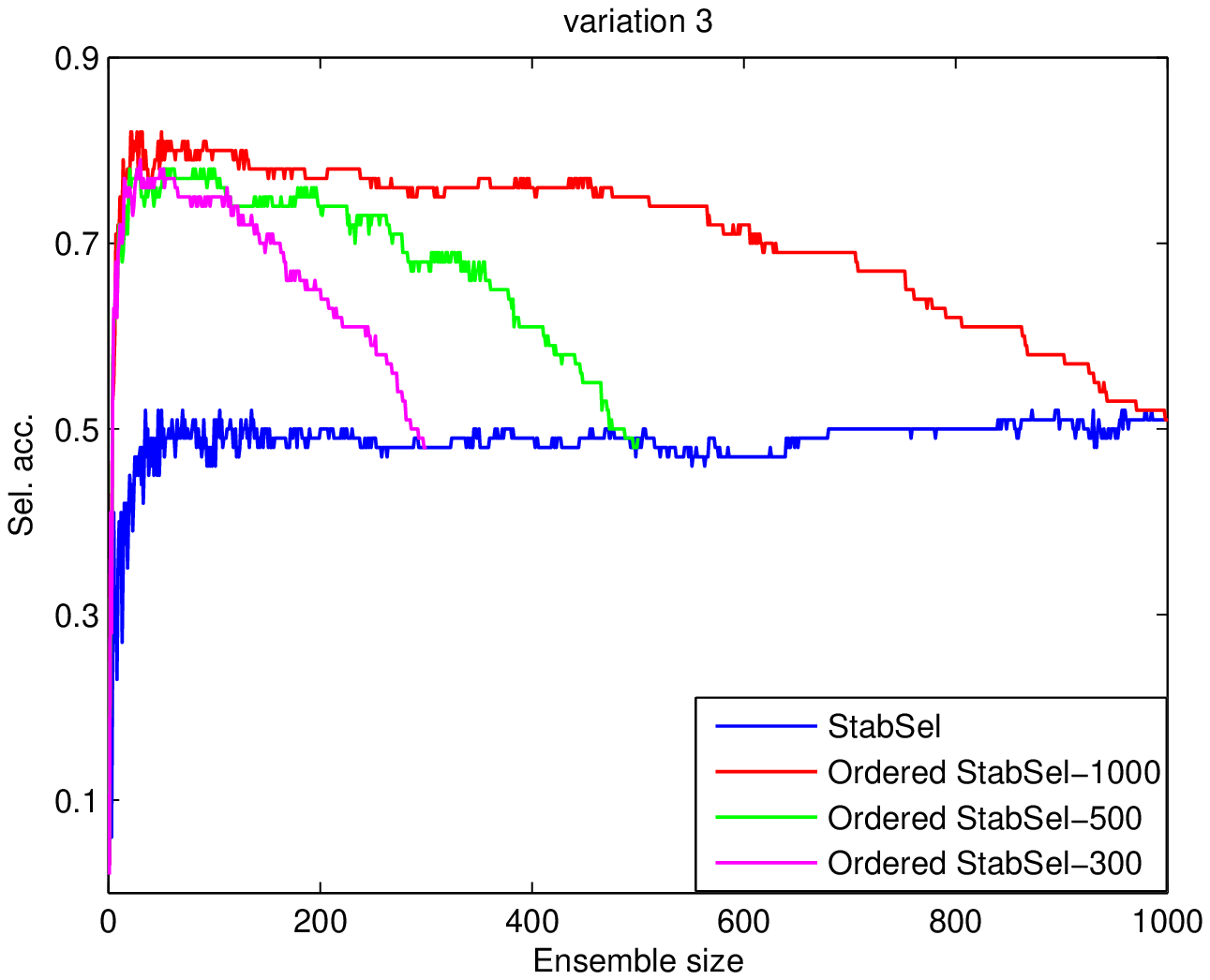}}}
\end{minipage}%
\hspace{0.005\textwidth}%
\begin{minipage}[b]{0.49\textwidth}
\centering{\scalebox{0.55}[0.50]{\includegraphics{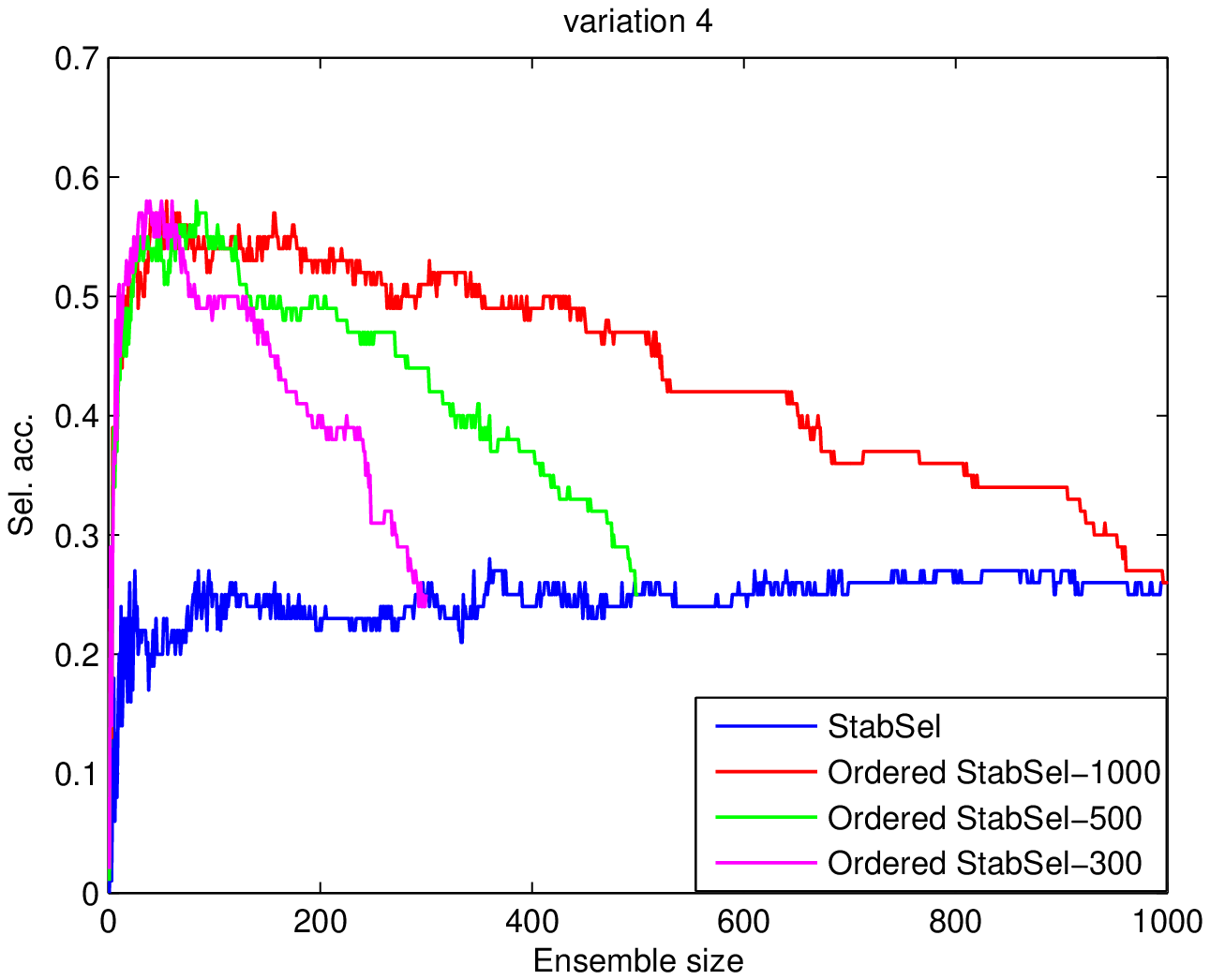}}}
\end{minipage}\\
\caption{The performance of StabSel and different sizes of ordered StabSel as a function of ensemble size. \label{fig:2}}
\end{figure}

There is some evidence in the literature of selective ensemble learning for classification that increasing the size of the initial pool of classifiers (i.e., the ensemble size $B$) generally improved the performance of the final pruned ensemble~\citep{Martinez2009}. To verify whether this was true for pruned VSEs as well, we used scenario 1 again to conduct the following experiments. For each variation of scenario 1, an initial StabSel ensemble of size 1000 was built. The ensemble members were then ordered, considering only the first 300, 500 and 1000 individuals of the original ensemble, respectively. Similar to the experiments of the previous section, these steps were repeated 100 times to estimate the respective selection accuracies of the full and pruned ensembles. The accuracy curves in Figure \ref{fig:2} illustrate that the maximum selection accuracies achieved by re-ordering an initial pool of $B = 300,\ 500$ or 1000 base learners are almost the same.

\subsection{Performance comparisons}
\label{section:5.4}

We now proceed to general comparisons. Based on the simulations in Sections~\ref{section:5.2} and \ref{section:5.3}, we used an ensemble size of $B=100$ in all subsequent experiments to strike a reasonable balance between performance and computational cost. Furthermore, these simulations (see Figures \ref{fig:1} and \ref{fig:2}) provided overwhelming evidence that only keeping a relatively small number of top-ranked ensemble members (often less than half) was usually sufficient to produce a better subensemble. As a result, in all experiments below we kept only the top $1/3$ after re-ordering the initial StabSel ensemble to form our pruned ensemble. To evaluate the performance metrics, every simulation was repeated for $M=500$ times when $p\leq n$ and $M=200$ times when $p>n$.

\begin{table}
\caption{Performance comparison for scenario 2, compound symmetric covariance structure. \label{tab:1}}
\begin{center}\vskip -0.5cm
{\small{\begin{tabular}{ccclcrrrrr}
\hline
$n$ & $p$ & $\rho$& Method & ~ &
\multicolumn{1}{c}{$\bar{p}_0$} &
\multicolumn{1}{c}{$\bar{p}_1$} &
\multicolumn{1}{c}{$acc.$} &
\multicolumn{1}{c}{FDR} &
\multicolumn{1}{c}{PErr(std)} \\
\hline
100 & 50 & 0 & StabSel     & & 0.021 & 0.997 & 0.332 & 0.145 & 0.118(0.084)\\
& &  & Pr. StabSel & & 0.009 & 0.992 & 0.612 & 0.067 & 0.094(0.082)\\
& &  & Lasso       & & 0.058 & 0.996 & 0.224 & 0.274 & 0.181(0.144)\\
& &  & SCAD        & & 0.048 & 0.971 & 0.184 & 0.241 & 0.185(0.130)\\
& &  & SIS         & & 0.049 & 0.924 & 0.128 & 0.251 & 0.290(0.317)\\
\cline{3-10}
& & 0.5 & StabSel      & & 0.007 & 0.943 & 0.528 & 0.058 & 0.115(0.090)\\
& & & Pr. StabSel  & & 0.002 & 0.912 & 0.504 & 0.022 & 0.127(0.089)\\
& & & Lasso   & & 0.132 & 0.997 & 0.004 & 0.514 & 0.175(0.094)\\
& & & SCAD         & & 0.048 & 0.971 & 0.184 & 0.241 & 0.185(0.129)\\
& & & SIS          & & 0.036 & 0.911 & 0.230 & 0.192 & 0.237(0.252)\\
\hline
100 & 100 & 0 & StabSel     & & 0.006 & 0.977 & 0.498 & 0.088 & 0.125(0.112)\\
& & & Pr. StabSel & & 0.002 & 0.952 & 0.630 & 0.033 & 0.132(0.123)\\
& & & Lasso& & 0.042 & 0.993 & 0.138 & 0.347 & 0.241(0.199)\\
& &  & SCAD        & & 0.085 & 0.998 & 0.152 & 0.354 & 0.225(0.154)\\
& &  & SIS         & & 0.028 & 0.874 & 0.058 & 0.283 & 0.461(0.538)\\
\cline{3-10}
& & 0.5 & StabSel      & & 0.006 & 0.929 & 0.370 & 0.091 & 0.141(0.095)\\
& & & Pr. StabSel  & & 0.002 & 0.898 & 0.410 & 0.031 & 0.143(0.089)\\
 & & & Lasso   & & 0.091 & 0.992 &$<0.001$ & 0.605 & 0.233(0.136)\\
& & & SCAD         & & 0.004 & 0.927 & 0.430 & 0.063 & 0.136(0.091)\\
& & & SIS          & & 0.023 & 0.837 & 0.114 & 0.252 & 0.396(0.419)\\
\hline
200 & 1000 & 0 & StabSel     & & $<0.001$ & 0.999 & 0.605 & 0.071 & 0.054(0.043)\\
& &  & Pr. StabSel & & $<0.001$ & 0.991 & 0.870 & 0.016 & 0.044(0.059)\\
& & & Lasso       & & 0.005 & 1.000 & 0.200 & 0.351 & 0.169(0.138)\\
& &  & SCAD        & & 0.012 & 0.999 & 0.085 & 0.536 & 0.273(0.163)\\
& &  & SIS         & & 0.005 & 0.864 & 0.060 & 0.375 & 0.381(0.358)\\
\cline{3-10}
& & 0.5 & StabSel      & & 0.001 & 0.964 & 0.420 & 0.095 & 0.080(0.067)\\
& & & Pr. StabSel  & & $<0.001$ & 0.942 & 0.590 & 0.031 & 0.078(0.074)\\
& & & Lasso   & & 0.020 & 0.997 & $<0.001$ & 0.790 & 0.212(0.072)\\
& & & SCAD         & & 0.018 & 0.996 & $<0.001$ & 0.770 & 0.205(0.079)\\
& & & SIS          & & 0.004 &  0.807 & 0.115 & 0.366 & 0.380(0.377)\\
\end{tabular}}}
\end{center}
\end{table}

In scenario 2, for each $\rho=0, 0.5$, we compared all methods when $n\geq p$, specifically, $(n,p)=(100,50),(100,100)$, and when $n \ll p$, specifically, $(n,p)=(200,1000)$. Table \ref{tab:1} reports the performance of each method, as measured by different metrics. In each row, the number in the parentheses of the last column is the standard error of $PErr$ from $M$ simulations. The following observations can be made. Firstly, all methods could detect important variables in most cases, as shown by $\bar{p}_1$, with the performance of SIS being slightly worse than others, especially when $p\geq n$. However, the lasso, SCAD and SIS all paid additional prices by including a relatively large number of uninformative variables, as indicated by the metrics $\bar{p}_0$, $acc.$ and FDR. Secondly, StabSel and pruned StabSel behaved significantly better than their rivals in terms of all metrics. Their advantages were more prominent in terms of $acc.$ and FDR. More importantly, pruned StabSel often significantly improved upon StabSel in being able to correctly identify the true model; in some cases, the selection accuracy ($acc.$) almost doubled. Thirdly, the prediction abilities of StabSel and of pruned StabSel were comparable, and both outperformed the other benchmark algorithms.
Here, the advantage of using a VSE over a single selector can also be seen clearly by comparing StabSel or pruned StabSel with the lasso, SCAD and SIS.

Table 2 summarizes the results for scenario 3. In this situation, it was difficult for any method to distinguish informative variables from uninformative ones because of their high correlation. The results in Table 2 show that the lasso, SCAD and SIS were almost useless in this case; their selection accuracy was almost zero and their FDRs were also very high. Because SIS utilizes the correlation between each covariate with the response to achieve variable screening, the spurious correlations in this scenario caused it to behave badly. By contrast, StabSel and pruned StabSel performed much better. Moreover, by eliminating some unnecessary members in the StabSel ensemble, the pruned StabSel ensemble was able to reach much better selection results (much higher accuracy and lower FDR).

\begin{table}
\caption{Performance comparison for scenario 3, block covariance structure. \label{tab:2}}
\begin{center}
{\small{\begin{tabular}{lcrrrrr}
\hline
Method & &
\multicolumn{1}{c}{$\bar{p}_0$} &
\multicolumn{1}{c}{$\bar{p}_1$} &
\multicolumn{1}{c}{$acc.$} &
\multicolumn{1}{c}{FDR} &
\multicolumn{1}{c}{PErr(std)} \\
\hline
Stabsel     & & 0.001 & 0.957 & 0.365 & 0.110 & 0.093(0.090)\\
Pr. StabSel & & $<0.001$ & 0.945 & 0.565 & 0.034 & 0.092(0.097)\\
Lasso   & & 0.019 & 0.982 & $<0.001$ & 0.782 & 0.130(0.099)\\
SCAD        & & 0.019 & 0.967 & $<0.001$ & 0.792 & 0.120(0.072)\\
SIS         & & 0.013 & 0.026 &	$<0.001$ & 0.987 & 5.992(0.869)\\
\end{tabular}}}
\end{center}
\end{table}

Results for scenario 4 are reported in Table 3. We can observe that StabSel and pruned StabSel both had satisfactory performances even when $\rho=0.90$, with pruned StabSel again significantly outperforming (regular) StabSel in terms of $acc.$ and FDR. However, the other methods could hardly detect the true model at all due to the two weak signals.

\begin{table}[!htbp]
\caption{Performance comparison for scenario 4, weak signals. \label{tab:3}}
\begin{center}
{\small{\begin{tabular}{clcrrrrr}
\hline
$\rho$& Method & &
\multicolumn{1}{c}{$\bar{p}_0$} &
\multicolumn{1}{c}{$\bar{p}_1$} &
\multicolumn{1}{c}{$acc.$} &
\multicolumn{1}{c}{FDR} &
\multicolumn{1}{c}{PErr(std)} \\
\cline{4-8}
\hline
& Stabsel     & & $<0.001$ & 1.000 & 0.675 & 0.062 & 0.047(0.038)\\
& Pr. StabSel & & $<0.001$ & 0.998 & 0.890 & 0.017 & 0.034(0.030)\\
0.50& Lasso   & & 0.004 & 0.629 & $<0.001$ & 0.478 & 0.094(0.113)\\
& SCAD        & & 0.004 & 0.990 & 0.160 & 0.347 & 0.174(0.117)\\
& SIS         & & 0.005 & 0.988 & 0.040 & 0.443 & 0.139(0.072)\\
\hline
& Stabsel     & & 0.001 & 0.983 & 0.340 & 0.133 & 0.043(0.031)\\
& Pr. StabSel & & $<0.001$ & 0.939 & 0.500 & 0.057 & 0.043(0.035)\\
0.90& Lasso   & & 0.002 & 0.750 & $<0.001$ & 0.382 &	0.046(0.044)\\
& SCAD        & & $<0.001$ & 0.438 & $<0.001$ & 0.067 & 0.560(0.127)\\
& SIS         & & 0.003 & 0.804 & $<0.001$ & 0.288 & 0.099(0.060)\\
\end{tabular}}}
\end{center}
\end{table}

Finally, Table 4 shows the results for scenario 5, a logistic regression problem, for both $p<n$ and $p>n$. From Table 4, we can draw some similar conclusions. The pruned StabSel ensemble continued to maintain its superiority over the other methods, especially in terms of selection accuracy and the FDR. 

The simulation results presented in this section strongly indicate that pruned StabSel is a competitive tool for performing variable selection in high-dimensional sparse models. As far as our current study is concerned, the most important message is that our ordering-based pruning algorithm (Algorithm 1) can give VSE algorithms such as StabSel a significant performance boost.

\begin{table}[!htbp]
\caption{Performance comparison for scenario 5, logistic model. \label{tab:4}}
\begin{center}
{\small{\begin{tabular}{clcrrrrr}
\hline
$p$ & Method & &
\multicolumn{1}{c}{$\bar{p}_0$} &
\multicolumn{1}{c}{$\bar{p}_1$} &
\multicolumn{1}{c}{$acc.$} &
\multicolumn{1}{c}{FDR} &
\multicolumn{1}{c}{PErr(std)} \\
\hline
& Stabsel     & & 0.014 & 1.000 & 0.522 & 0.146 & 0.123(0.024)\\
& Pr. StabSel & & 0.006 & 0.999 & 0.744 & 0.069 & 0.121(0.024)\\
50 & Lasso    & & 0.054 & 1.000 & 0.228 & 0.353 & 0.129(0.013)\\
& SCAD        & & 0.087 & 0.999 & 0.036 & 0.516 & 0.136(0.011)\\
& SIS         & & 0.084 & 0.999 & 0.044 & 0.514 & 0.135(0.011)\\
\hline
& Stabsel     & & $<0.001$ & 0.993 & 0.790 & 0.051 & 0.120(0.024)\\
& Pr. StabSel & & $<0.001$ & 0.980 & 0.900 & 0.010 & 0.119(0.023)\\
1000 & Lasso  & & 0.009 & 1.000 &   0.115  & 0.565 & 0.159(0.031)\\
& SCAD        & & 0.020 & 0.997 & $<0.001$ & 0.853 & 0.180(0.021)\\
& SIS         & & 0.012 & 0.997 & $<0.001$ & 0.794 & 0.179(0.023)
\end{tabular}}}
\end{center}
\end{table}

\section{Real Data Analysis}
\label{section:6}

To assess how well each method behaves on real data, we took two real data sets and followed an evaluation procedure utilized by other researchers \citep{Buhlmann2014b,Lin2016}. In particular, the design matrix ${\bf X}$ of the real data set was used with randomly generated coefficients and error terms to produce the response, so one knew beforehand whether each variable was important or not. Our first example is the Riboflavin data set from \citet{Buhlmann2014a}. It is for a regression task with 111 observations and 4088 continuous covariates. Our second example is the Madelon data set from the UCI repository \citep{Lichman2013}. It is for a binary classification problem, which has been used as part of the NIPS 2003 feature selection challenge. There are 2600 observations and 500 variables.

For the Riboflavin data, we first drew $p$ variables at random. Next, the number of nonzero coefficients was set to be $s$ and their true values were randomly taken to be $1$ or $-1$. Then, responses were created by adding error terms generated from a normal distribution $N(0,\sigma^2)$, where $\sigma^2$ was determined to achieve a specific signal-to-noise ratio ($snr$). Finally, to evaluate the predictive performance of the selected models, these data were randomly split into a training set (90\%) and a test set (10\%). For any method under investigation, we first applied it to the training set to perform variable selection. Based on the selected variables, we then built a linear regression model and estimated its prediction error on the test set. The entire process was repeated 200 times. For the Madelon data, a similar process was followed, except that, instead of adding normally distributed error terms to generate the responses, we simply generated each response $y_i$ from a binomial distribution with probability $1/[1 + \exp(-{\bf x}_i^{\rm T}\boldsymbol{\beta})]$. Since we were mostly interested in the behavior of each method in relatively high-dimensional (large $p$) situations, only 400 observations were used for training and the remaining ones were taken as the test set. Tuning parameters for each method --- such as $\pi_{\rm thr}$ and $\lambda_{\rm min}$ for StabSel --- were specified in the same manner as they were in the simulation studies (see Sections~\ref{section:5} and \ref{section:6}).

\begin{table}[!t]
\caption{Performance comparison on real-data examples. \label{tab:5}}
\begin{center}
{\small{\begin{tabular}{ccclccrrrrr}
\hline
Dataset & $(s,p)$ & $snr$ & Method & &
\multicolumn{1}{c}{$\bar{p}_0$} &
\multicolumn{1}{c}{$\bar{p}_1$} &
\multicolumn{1}{c}{$acc.$} &
\multicolumn{1}{c}{FDR} &
\multicolumn{1}{c}{PErr(std)} \\
\hline
& & & StabSel     & & 0.012 & 0.745 & 0.120 & 0.227 & 2.422(1.714)\\
& & & Pr. StabSel & & 0.004 & 0.691 & 0.170 & 0.113 & 2.649(2.010)\\
Riboflavin & $(5,100)$ & 3 &Lasso& & 0.083 & 0.906 &$<0.001$ &0.592 & 2.391(1.612)\\
& & & SCAD        & & 0.045 & 0.922 & 0.100 & 0.424 & 2.170(1.578)\\
& & & SIS         & & 0.051 & 0.556 & 0.005 & 0.563 & 3.128(1.900)\\
\hline
& & & StabSel     & & 0.006 & 0.837 & 0.140 & 0.201 & 1.017(0.754)\\
& & & Pr. StabSel & & 0.003 & 0.810 & 0.245 & 0.121 & 1.065(0.777)\\
Riboflavin & $(5,200)$ & 8 & Lasso  & & 0.065 & 0.980 & $<0.001$& 0.694& 0.888(0.546)\\
& & & SCAD        & & 0.092 & 0.993 & $<0.001$ & 0.773 & 0.920(0.655)\\
& & & SIS         & & 0.033 & 0.508 & 0.005 & 0.640 & 1.898(1.198)\\
\hline
& & & StabSel     & & 0.004 & 0.989 & 0.395 & 0.069 & 0.171(0.012)\\
& & & Pr. StabSel & & 0.002 & 0.985 & 0.590 & 0.039 & 0.170(0.014)\\
Madelon & $(10,200)$ & / & Lasso & & 0.080 & 0.996 & $<0.001$ & 0.560 & 0.201(0.017)\\
& & & SCAD        & & 0.084 & 0.988 & $<0.001$ & 0.597 & 0.199(0.015)\\
& & & SIS         & & 0.078 & 0.988 & $<0.001$ & 0.583 & 0.198(0.014)\\
\end{tabular}}}
\end{center}
\end{table}

Table 5 summarizes the results. It can be seen that the pruned StabSel ensemble again achieved the best performance in all cases as measured by $\bar{p}_0$, $acc.$ and FDR. When the ratio $snr$ was high and the model was sparse (small $s$ relative to $p$), its relative advantage over other methods was more prominent. Although pruned StabSel generally has a slightly lower true positive rate ($\bar{p}_1$) --- a necessary consequence of having a much reduced false positive rate ($\bar{p}_0$), its overall selection accuracy tends to be much higher than other methods. Finally in terms of prediction capacity, pruned StabSel is better than or competitive with other algorithms.

\section{Conclusions}
\label{section:7}

In this paper, we have investigated the idea of selective ensemble learning for constructing VSEs. In particular, we have developed a novel ordering-based ensemble pruning technique to improve the selection accuracy of VSEs. By rearranging aggregation order of the ensemble members, we can construct a subensemble by fusing only members ranked at the top. More specifically, each member is sequentially included into the ensemble so that at each step the loss between the resulting ensemble's importance vector and a reference vector is minimized. This novel technique can be applied to any VSE algorithm, but in our experiments with both simulated and real-world data, we have largely focused on using it to boost the performance of stability selection (StabSel), a particular VSE technique, partly because of the latter's popularity and flexibility but also because it is not directly obvious how our technique can be applied as StabSel does not aggregate information from its members with simple averaging. Our empirical results have been overwhelmingly positive. As such, a pruned StabSel ensemble can be considered an effective alternative to perform variable selection in real applications, especially those with high dimensionality.

%
%
%
%
%

%

\bibliographystyle{Chicago}

\bibliography{Bibliography-MM-MC}

\end{document}